\documentclass[11pt, a4paper]{article}

\usepackage[a4paper, top=2.5cm, bottom=2.5cm, left=2.5cm, right=2.5cm]{geometry}
\usepackage{amsmath, amssymb, amsfonts}
\usepackage{booktabs}
\usepackage{multirow}
\usepackage{authblk}
\usepackage{abstract}
\usepackage{hyperref}
\usepackage{cleveref}
\usepackage{enumitem}
\usepackage{graphicx}
\usepackage{pifont}
\usepackage{mathptmx}
\usepackage{geometry}
\usepackage{authblk}
\usepackage{mathptmx}
\usepackage{microtype} 
\usepackage{graphicx} 
\usepackage{algorithm}
\usepackage{algpseudocode} 
\newcommand{\vect}[1]{\mathbf{#1}}
\newcommand{\mat}[1]{\mathbf{#1}}
\newcommand{\set}[1]{\mathcal{#1}}
\newcommand{\reals}{\mathbb{R}}
\newcommand{\E}{\mathbb{E}}

\title{\Large \bfseries Semantic-Topological Graph Reasoning for Language-Guided Pulmonary Screening}
\author[1]{Chenyu Xue}
\author[2]{Yiran Liu}
\author[1]{Mian Zhou}
\author[1]{Jionglong Su}
\author[3]{Zhixiang Lu}
\affil[1]{Xi'an Jiaotong-Liverpool University, Jiangsu, China}

\affil[2]{University College London, London, United Kingdom}

\affil[3]{University of Liverpool, Liverpool, United Kingdom \authorcr 
\tt{Zhixiang@liverpool.ac.uk}}

\date{}

\begin{document}

\maketitle

\begin{abstract}
Medical image segmentation driven by free-text clinical instructions is a critical frontier in computer-aided diagnosis. However, existing multimodal and foundation models struggle with the semantic ambiguity of clinical reports and fail to disambiguate complex anatomical overlaps in low-contrast scans. Furthermore, fully fine-tuning these massive architectures on limited medical datasets invariably leads to severe overfitting. To address these challenges, we propose a novel Semantic-Topological Graph Reasoning (STGR) framework for language-guided pulmonary screening. Our approach elegantly synergizes the reasoning capabilities of large language models (LLaMA-3-V) with the zero-shot delineation of vision foundation models (MedSAM). Specifically, we introduce a Text-to-Vision Intent Distillation (TVID) module to extract precise diagnostic guidance. To resolve anatomical ambiguity, we formulate mask selection as a dynamic graph reasoning problem, where candidate lesions are modeled as nodes and edges capture spatial and semantic affinities. To ensure deployment feasibility, we introduce a Selective Asymmetric Fine-Tuning (SAFT) strategy that updates less than 1\% of the parameters. Rigorous 5-fold cross-validation on the LIDC-IDRI and LNDb datasets demonstrates that our framework establishes a new state-of-the-art. Notably, it achieves an 81.5\% Dice Similarity Coefficient (DSC) on LIDC-IDRI, outperforming leading LLM-based tools like LISA by over 5\%. Crucially, our SAFT strategy acts as a powerful regularizer, yielding exceptional cross-fold stability ($\pm 0.6\%$ DSC variance) and paving the way for robust, context-aware clinical deployment.
\end{abstract}

\vspace{2mm}

\section{Introduction}
\label{sec:intro}
Referring Expression Segmentation (RES) adapted for the medical domain aims to localize and segment pathological lesions based on free-form clinical texts or radiologist reports. As computer-aided diagnosis (CAD) workflows rapidly evolve to handle high-throughput screening environments, there is an escalating demand for intelligent systems capable of interpreting nuanced clinical instructions (e.g., ``diffuse ground-glass opacities predominantly in the upper right lobe''). While specialized CAD systems have demonstrated high interpretability in targeted rapid screening tasks such as lung nodule segmentation \cite{hu2024lung}, conventional segmentation networks like U-Net \cite{ronneberger2015u} rely strictly on fixed label spaces. This renders them intrinsically incapable of adapting to the arbitrary and variable nature of language-driven diagnostic prompts.

Recent breakthroughs in Large Multimodal Models (LMMs) and Vision Foundation Models have catalyzed new paradigms for vision-language integration in healthcare, showcasing immense potential in cross-modal alignment for tasks like dermatological diagnosis \cite{Lu2026skinclipvl}. However, adapting models such as LLaMA-3-V \cite{llama3} or the medical Segment Anything Model (MedSAM) \cite{ma2024segment} directly to pulmonary RES introduces significant bottlenecks. LMMs fundamentally lack the pixel-level spatial acuity required for fine-grained boundary delineation. Conversely, while MedSAM possesses robust zero-shot capabilities, it requires highly deterministic geometric prompts (e.g., exact bounding boxes) and frequently fails when confronted with the semantic ambiguity of overlapping tissues on low-contrast radiographs. While recent efforts have explored utilizing LLMs as causal reasoners for medical segmentation \cite{Causalsamllm}, these approaches often overlook the critical topological interactions between adjacent anatomical structures.

To bridge this gap, we propose a novel multimodal framework specifically tailored for complex pulmonary screening. Our architecture elegantly synergizes the semantic comprehension of LMMs, the zero-shot delineation of foundation models, and the topological reasoning capabilities of graph networks. Our primary contributions are threefold:
\begin{itemize}[leftmargin=*, noitemsep]
    \item \textbf{Text-to-Vision Intent Distillation:} We introduce an instruction-decomposition strategy utilizing an adapted LLaMA-3-V model to extract rich, pathology-specific semantic vectors from complex clinical texts, directly guiding the downstream visual cascade.
    \item \textbf{Semantic-Topological Graph Reasoning (STGR):} By modeling the topological and semantic affinities between candidate lesion masks, our network performs collaborative graph reasoning to disambiguate target lesions from overlapping normal tissues, solving a critical failure mode of standard linear projection heads.
    \item \textbf{Ultra-Efficient Clinical Adaptation:} We design a Selective Asymmetric Fine-Tuning (SAFT) strategy. By freezing all heavy backbones and exclusively training lightweight Adapters and low-rank matrices, we reduce trainable parameters to $\sim 0.6\%$, democratizing clinical model deployment.
\end{itemize}

\section{Related Work}
\label{sec:related}

\subsection{Multimodal Medical AI and Vision-Language Alignment}
The integration of textual data with medical imagery has significantly advanced diagnostic workflows. Beyond standard image-only networks \cite{chen2021transunet}, recent methodologies have heavily emphasized cross-modal consistency. For example, Lu et al. \cite{Lu2026skinclipvl} demonstrated the efficacy of consistency-aware vision-language learning (SkinCLIP-VL) for multimodal diagnostic alignment, while highly optimized gradient-boosted convolutional pipelines have pushed the boundaries of rapid, interpretable clinical screening \cite{DeepGBTB2026}. Expanding upon these vision-language alignments, our work utilizes Foundation Models (e.g., MedSAM \cite{ma2024segment} and GroundingDINO \cite{liu2023grounding}) to generate candidate region proposals, directly supervised by language-driven embeddings.

\subsection{LLM Reasoning and Parameter-Efficient Tuning Paradigms}
The deployment of LLMs as central reasoning engines is transforming segmentation tasks \cite{lai2023lisa, zou2023segment}. Recent advances have successfully positioned LLMs as causal reasoners to enforce logical robustness in medical segmentation pipelines \cite{Causalsamllm}. Concurrently, the broader NLP community has developed highly sophisticated optimization frameworks for LLMs in complex, low-resource environments. Techniques ranging from agent-guided expert-tuning \cite{Lu2026sage,lu2026merit} and unified data optimization strategies \cite{Lu2025advancing} to personality-driven multi-agent simulations \cite{Lu2025prism} have proven that LLMs can deeply understand highly specific, nuanced instructions when tuned efficiently. Inspired by these NLP optimization paradigms, our Selective Asymmetric Fine-Tuning (SAFT) and Text-to-Vision Intent Distillation (TVID) modules leverage low-rank adaptation \cite{dettmers2023qlora} to affordably parse complex radiological prompts.

\subsection{Complex Feature Mining and Structural Reasoning}
Resolving anatomical ambiguity in medical scans requires advanced structural modeling beyond simple convolutional receptive fields. The data mining community has consistently demonstrated that structured attention and hierarchical modeling are critical for representing complex, noisy relationships. This is evident in diverse applications, from attention-based deep learning frameworks for material compressive strength \cite{Lu2026attention}, to hierarchical risk prediction on social networks \cite{Lu2025hierrisk}, and dynamic identity equilibrium strategies for complex entity disambiguation \cite{Lu2024dieq}. Drawing a parallel to these high-dimensional data mining successes, our Semantic-Topological Graph Reasoning (STGR) module treats candidate masks as a complex network of entities, leveraging spatial and semantic edges to resolve structural ambiguities in pulmonary screening.

\begin{figure}[t!]
  \centering
    \includegraphics[width=\textwidth]{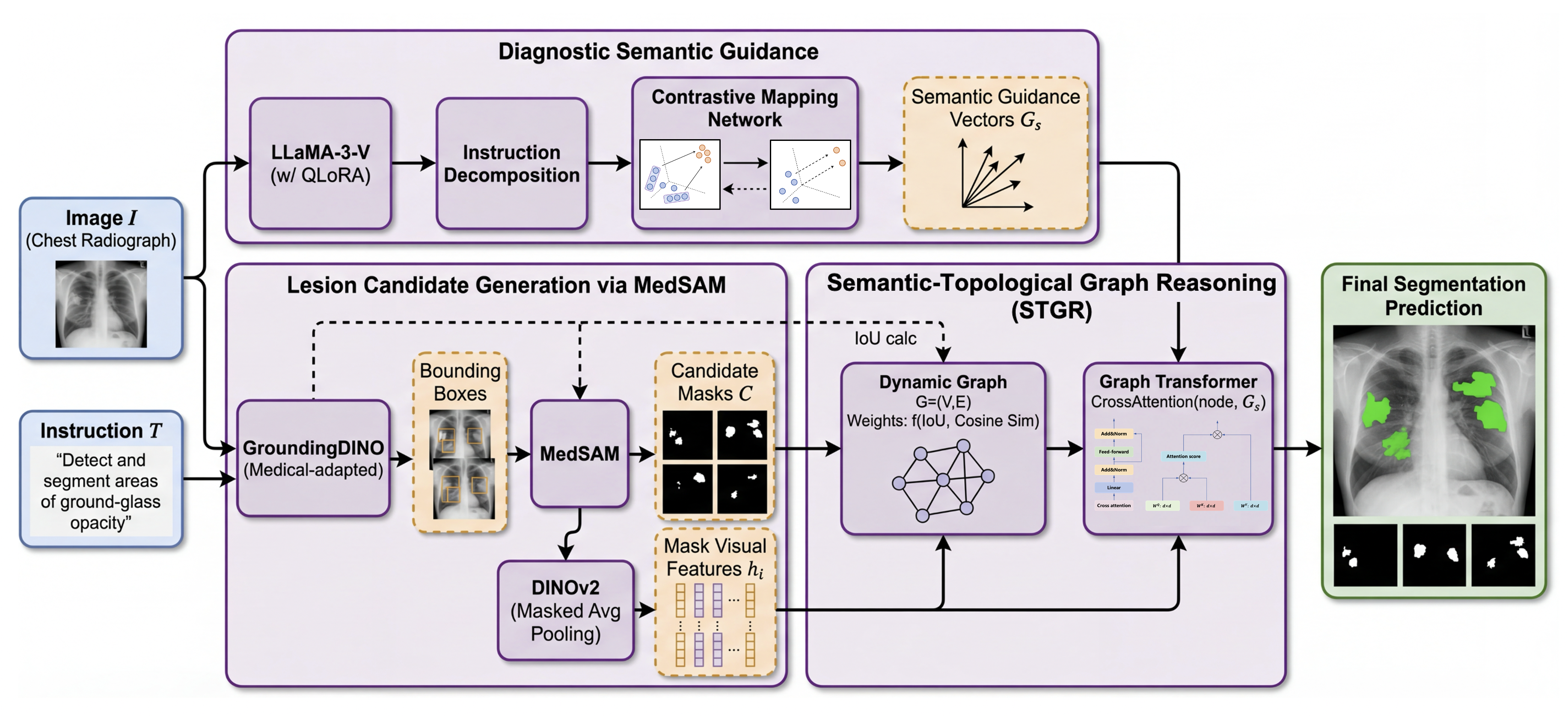}
  \caption{The overall framework of our proposed language-guided pulmonary segmentation model. The framework consists of three stages: (A) LLaMA-3-V-driven Diagnostic Semantic Generation, (B) MedSAM-based Candidate Mask Generation and (C) Graph-Structured Collaborative Reasoning.}
  \label{fig:pipeline}
\end{figure}

\section{Proposed Methodology}
\label{sec:method}
To clearly illustrate the forward pass of our Semantic-Topological Graph Reasoning (STGR) framework, we detail the core execution steps in Algorithm \ref{alg:stgr}. The framework seamlessly integrates the rich linguistic priors of LLaMA-3-V with the precise spatial localization of MedSAM, performing collaborative reasoning across candidate pathological regions.

\begin{algorithm}[t!]
\caption{Forward Pass of the STGR Framework}
\label{alg:stgr}
\begin{algorithmic}[1]
\Require Chest Radiograph $\mathbf{I}$, Clinical Instruction $T$, Graph Layers $L$, Threshold $\tau$
\Ensure Final Pathological Segmentation Mask $M_{final}$

\Statex \textbf{\textit{Phase 1: Intent Distillation \& Candidate Generation}}
\State $G_s \gets \text{LLaMA-3-V-TVID}(T)$ \Comment{Extract multimodal semantic guidance vectors}
\State $\{b_j\} \gets \text{GroundingDINO}(\mathbf{I}, T)$ \Comment{Generate open-set bounding boxes}
\State $\mathcal{C} \gets \text{MedSAM}(\mathbf{I}, \{b_j\})$ \Comment{Generate candidate masks $\{m_1, \dots, m_N\}$}
\For{each $m_i \in \mathcal{C}$}
    \State $\mathbf{h}_i \gets \text{DINOv2}(\mathbf{I} \odot m_i)$ \Comment{Extract visual features via masked pooling}
\EndFor

\Statex \textbf{\textit{Phase 2: Dynamic Graph Construction}}
\State Initialize graph nodes $V = \{\mathbf{h}_1, \dots, \mathbf{h}_N\}$
\For{each pair $(m_i, m_j) \in \mathcal{C} \times \mathcal{C}$}
    \State $e_{ij} \gets \alpha \cdot \text{IoU}(m_i, m_j) + \beta \cdot \text{CosineSim}(\mathbf{h}_i, \mathbf{h}_j)$ \Comment{Compute edge affinity}
\EndFor
\State Construct fully connected dynamic graph $\mathcal{G} = (V, E)$

\Statex \textbf{\textit{Phase 3: Semantic-Topological Reasoning}}
\For{$l = 1$ to $L$}
    \For{each node $v_i \in V$}
        \State $\mathbf{h}_{i, \text{attn}}^{(l)} \gets \text{SelfAttention}(\mathbf{h}_i^{(l)}, \mathcal{N}(v_i))$ \Comment{Spatial \& topological aggregation}
        \State $\mathbf{h}_i^{(l+1)} \gets \text{CrossAttention}(\mathbf{h}_{i, \text{attn}}^{(l)}, G_s)$ \Comment{Diagnostic semantic injection}
    \EndFor
\EndFor

\Statex \textbf{\textit{Phase 4: Confident Mask Selection}}
\State $\mathcal{C}_{selected} \gets \emptyset$
\For{each node $v_i \in V$}
    \If{$\text{MLP-Classifier}(\mathbf{h}_i^{(L+1)}) > \tau$}
        \State $\mathcal{C}_{selected} \gets \mathcal{C}_{selected} \cup \{m_i\}$
    \EndIf
\EndFor
\State $M_{final} \gets \bigcup_{m \in \mathcal{C}_{selected}} m$ \Comment{Merge highly confident masks}
\State \Return $M_{final}$
\end{algorithmic}
\end{algorithm}

\subsection{Text-to-Vision Intent Distillation (TVID)}
The first module distills high-level diagnostic intent from a radiological scan $\mat{I} \in \reals^{H \times W \times 3}$ and a corresponding clinical text instruction $T$. We employ an instruction decomposition strategy where our adapted LLaMA-3-V parses $T$ into $K$ distinct pathological attributes. For each attribute, the model outputs a latent text vector, forming a set $V_s$. To bridge the modality gap, these vectors are projected into the visual feature space via a non-linear mapping network $\Phi_{\text{proj}}$, yielding aligned semantic guidance vectors $G_s = \{\vect{g}_s^{(k)}\}_{k=1}^K$:
\begin{equation}
\vect{g}_s^{(k)} = \Phi_{\text{proj}}\left(\vect{v}_s^{(k)}; \Theta_{\text{proj}}\right) \in \reals^{d_v}
\label{eq:mapping}
\end{equation}
where $d_v$ denotes the dimensionality of the visual embedding space.

\subsection{Lesion Candidate Generation via Foundation Models}
To obtain accurate region boundaries without relying on extensive domain-specific dense annotations, we utilize a cascaded zero-shot approach. Initially, a medical-adapted GroundingDINO \cite{liu2023grounding} detects preliminary regions of interest based on the textual prompt $T$, outputting a set of bounding boxes. These boxes act as spatial prompts for MedSAM \cite{ma2024segment}, generating a dense pool of high-quality, albeit occasionally noisy or overlapping, candidate masks $\set{C} = \{m_1, \dots, m_N\}$. Subsequently, a frozen DINOv2 \cite{oquab2023dinov2} encoder extracts a multi-scale semantic feature vector $\vect{h}_i \in \reals^{d_v}$ for each candidate mask via masked average pooling.

\subsection{Semantic-Topological Graph Reasoning (STGR)}
The core challenge in medical RES is resolving anatomical ambiguity. A linear classifier often fails to distinguish a true pulmonary nodule from an overlapping vessel or rib structure due to visually similar localized textures. We resolve this by constructing a dynamic reasoning graph $\set{G} = (\set{V}, \set{E})$. Each node $v_i \in \set{V}$ is initialized with the candidate feature $\vect{h}_i$. 

The edge weights $\mathcal{E}(v_i, v_j)$ between nodes are explicitly formulated to capture both spatial intersection and morphological similarity:
\begin{equation}
\mathcal{E}(v_i, v_j) = \alpha \cdot \underbrace{\frac{|m_i \cap m_j|}{|m_i \cup m_j|}}_{\mathcal{S}_{\text{spatial}}} + (1 - \alpha) \cdot \underbrace{\frac{\langle \vect{h}_i, \vect{h}_j \rangle}{\|\vect{h}_i\|_2 \|\vect{h}_j\|_2}}_{\mathcal{S}_{\text{semantic}}}
\label{eq:edges}
\end{equation}
where $\alpha \in [0, 1]$ is a learnable balancing parameter. A Graph Transformer iteratively updates node representations to propagate context. Let $\vect{H}^{(l)}$ denote the node feature matrix at layer $l$. The message passing phase integrates neighborhood multi-head self-attention (MSA), followed by a cross-attention (MCA) mechanism conditioned on the diagnostic guidance vectors $G_s$:
\begin{equation}
\hat{\vect{H}}^{(l)} = \text{MSA}\left(\text{LayerNorm}(\vect{H}^{(l-1)}, \set{E})\right) + \vect{H}^{(l-1)}
\label{eq:graph_msa}
\end{equation}
\begin{equation}
\vect{H}^{(l)} = \text{MCA}\left(\text{LayerNorm}(\hat{\vect{H}}^{(l)}), G_s\right) + \hat{\vect{H}}^{(l)}
\label{eq:graph_mca}
\end{equation}
After $L$ layers, a Multi-Layer Perceptron (MLP) head outputs a final confidence score for each mask, retaining only the structures that possess both topological validity and strong semantic alignment with the clinical intent.

\subsection{Selective Asymmetric Fine-Tuning (SAFT)}
To prevent catastrophic forgetting and maintain computational tractability, we implement a Selective Asymmetric Fine-Tuning (SAFT) strategy. The backbones of LLaMA-3-V, GroundingDINO, MedSAM, and DINOv2 are strictly frozen. Trainable parameters are restricted to: (1) low-rank adaptation (QLoRA) \cite{dettmers2023qlora} matrices injected into LLaMA-3-V's attention layers to specialize its medical vocabulary, (2) the mapping network $\Phi_{\text{proj}}$, and (3) the STGR module.

The framework is optimized end-to-end using a composite loss function $\mathcal{L}_{\text{total}} = \mathcal{L}_{ce} + \lambda_1 \mathcal{L}_{\text{NCE}} + \lambda_2 \mathcal{L}_{reg}$. $\mathcal{L}_{ce}$ is the cross-entropy loss for mask selection. Critically, $\mathcal{L}_{\text{NCE}}$ is an InfoNCE-style contrastive loss that enforces proximity between the text embeddings $\vect{g}_s$ and the visual embeddings of true positive masks ($\vect{h}^+$), while repelling negative samples ($\mathcal{N}^-$):
\begin{equation}
\mathcal{L}_{\text{NCE}} = - \E_{k \sim \mathcal{K}} \left[ \log \frac{\exp \left( \langle \vect{g}_s^{(k)}, \vect{h}^+ \rangle / \tau \right)}{\exp \left( \langle \vect{g}_s^{(k)}, \vect{h}^+ \rangle / \tau \right) + \sum_{j \in \mathcal{N}^-} \exp \left( \langle \vect{g}_s^{(k)}, \vect{h}_j \rangle / \tau \right)} \right]
\end{equation}
where $\tau$ is a temperature scaling hyperparameter. Finally, $\mathcal{L}_{reg}$ utilizes Smooth $L_1$ loss to regress the predicted IoU quality of the candidates against the ground truth.

\section{Experiments \& Results}
\label{sec:experiments}

\textbf{Datasets and Evaluation Metrics.} We validated our framework on two challenging cohorts: the LIDC-IDRI dataset \cite{pehrson2019automatic} for fine-grained lung nodule detection, and a curated subset of LNDb \cite{pedrosa2019lndblungnoduledatabase} providing bounding box and mask annotations for broad pathological infiltrates. Following standard protocols, we report the Intersection over Union (IoU) and Dice Similarity Coefficient (DSC) across 5-fold cross-validation to ensure a rigorous and statistically sound evaluation. Let $P$ represent the predicted segmentation mask and $G$ represent the ground truth mask. The evaluation metrics are mathematically formulated as:
\begin{equation}
    \text{IoU} = \frac{|P \cap G|}{|P \cup G|}, \quad \text{DSC} = \frac{2|P \cap G|}{|P| + |G|}
\end{equation}

\textbf{Implementation Details.} Our framework is implemented in PyTorch and trained on a cluster of four NVIDIA A100 GPUs. All input radiographs are resized to $1024 \times 1024$ pixels and normalized to match the MedSAM processing requirements. We adopt the 8-billion parameter variant of LLaMA-3-V and a frozen DINOv2 (ViT-L/14) encoder for feature extraction. During the SAFT phase, we strictly freeze all foundation model backbones. For the language model, we apply QLoRA to the attention projection layers with a rank of $r=16$, a scaling factor $\alpha=32$, and a dropout rate of $0.05$. The Graph Transformer is configured with $L=3$ layers and bottleneck adapters of dimension $64$. The framework is optimized end-to-end using the AdamW optimizer with a batch size of 16 and an initial learning rate of $1 \times 10^{-4}$ decayed via a cosine annealing schedule over 50 epochs. The loss weighting coefficients are empirically set to $\lambda_{ce}=1.0$, $\lambda_{align}=0.5$, and $\lambda_{reg}=0.5$, with the final mask selection threshold $\tau$ set to $0.5$.

\textbf{Quantitative Results.} As summarized in Table \ref{tab:results}, our proposed framework (w/ SAFT) establishes a new state-of-the-art on both datasets. In the context of standard medical segmentation networks, our language-guided approach significantly reduces false positives by leveraging clinical context, outperforming strong baselines like SegNet (DSC 81.5\% vs. 74.3\% on LIDC-IDRI). Furthermore, when compared to leading vision foundation models and LLM-based segmentation tools like MedSAM and LISA, our method demonstrates a definitive advantage. Specifically, it surpasses LISA by a substantial margin of 5.3\% in DSC on LIDC-IDRI and 4.5\% on LNDb. Crucially, our model exhibits the lowest standard deviation across the 5-fold cross-validation ($\pm 0.6\%$ DSC on LIDC-IDRI), indicating exceptional robustness and stability across diverse patient cohorts, a vital requirement for clinical deployment.

\begin{table*}[t!]
\centering
\small
\renewcommand{\arraystretch}{1.25}
\setlength{\tabcolsep}{8pt} 
\caption{Quantitative evaluation on pulmonary screening datasets across 5-fold cross-validation. Results are reported as mean $\pm$ standard deviation. Best results are highlighted in \textbf{bold}.}
\label{tab:results}
\begin{tabular}{@{} l c c c c @{}}
\toprule
& \multicolumn{2}{c}{\textbf{LIDC-IDRI}} & \multicolumn{2}{c}{\textbf{LNDb}} \\
\cmidrule(lr){2-3} \cmidrule(lr){4-5}
\textbf{Method} & \textbf{IoU (\%)} & \textbf{DSC (\%)} & \textbf{IoU (\%)} & \textbf{DSC (\%)} \\
\midrule
\multicolumn{5}{@{}l}{\textit{\textbf{Standard Medical Segmentation Models}}} \\
U-Net \cite{ronneberger2015u}               & $54.2 \pm 1.2$ & $68.5 \pm 1.4$ & $49.3 \pm 1.1$ & $62.1 \pm 1.5$ \\
TransUNet \cite{chen2021transunet}         & $58.4 \pm 0.9$ & $72.1 \pm 1.1$ & $53.7 \pm 1.0$ & $66.8 \pm 1.2$ \\
SegNet \cite{segnet}                        & $61.2 \pm 1.1$ & $74.3 \pm 1.0$ & $55.1 \pm 0.8$ & $68.4 \pm 1.3$ \\
\midrule
\multicolumn{5}{@{}l}{\textit{\textbf{Foundation \& Referring Segmentation Models}}} \\
MedSAM (Text-Prompt) \cite{ma2024segment}   & $56.5 \pm 1.5$ & $70.3 \pm 1.6$ & $51.2 \pm 1.4$ & $65.0 \pm 1.7$ \\
SEEM~\cite{zou2023segment}                  & $59.8 \pm 1.0$ & $73.0 \pm 1.2$ & $54.9 \pm 0.9$ & $67.5 \pm 1.1$ \\
LISA~\cite{lai2023lisa}                     & $62.8 \pm 0.8$ & $76.2 \pm 0.9$ & $57.4 \pm 0.7$ & $70.1 \pm 0.9$ \\
\midrule
\multicolumn{5}{@{}l}{\textit{\textbf{Ablation \& Fine-Tuning Strategies (Ours)}}} \\
Baseline (LLaMA-3-V + MedSAM)                 & $59.5 \pm 1.2$ & $72.5 \pm 1.3$ & $55.2 \pm 1.1$ & $67.8 \pm 1.4$ \\
+ Text-to-Vision Intent Distillation (TVID) & $62.1 \pm 0.9$ & $75.3 \pm 1.0$ & $57.0 \pm 0.8$ & $69.4 \pm 1.1$ \\
+ Semantic-Topological Graph Reasoning (STGR)& $67.5 \pm 0.6$ & $80.1 \pm 0.7$ & $60.8 \pm 0.7$ & $73.2 \pm 0.8$ \\
Full Model (No PEFT, Full Tuning)           & $66.8 \pm 0.8$ & $79.5 \pm 0.9$ & $60.2 \pm 0.9$ & $72.8 \pm 1.0$ \\
\textbf{Ours (Full Model w/ SAFT)}          & $\mathbf{69.1 \pm 0.5}$ & $\mathbf{81.5 \pm 0.6}$ & $\mathbf{62.4 \pm 0.6}$ & $\mathbf{74.6 \pm 0.7}$ \\
\bottomrule
\end{tabular}
\end{table*}

\begin{figure*}[t!]
    \centering
    \includegraphics[width=\textwidth]{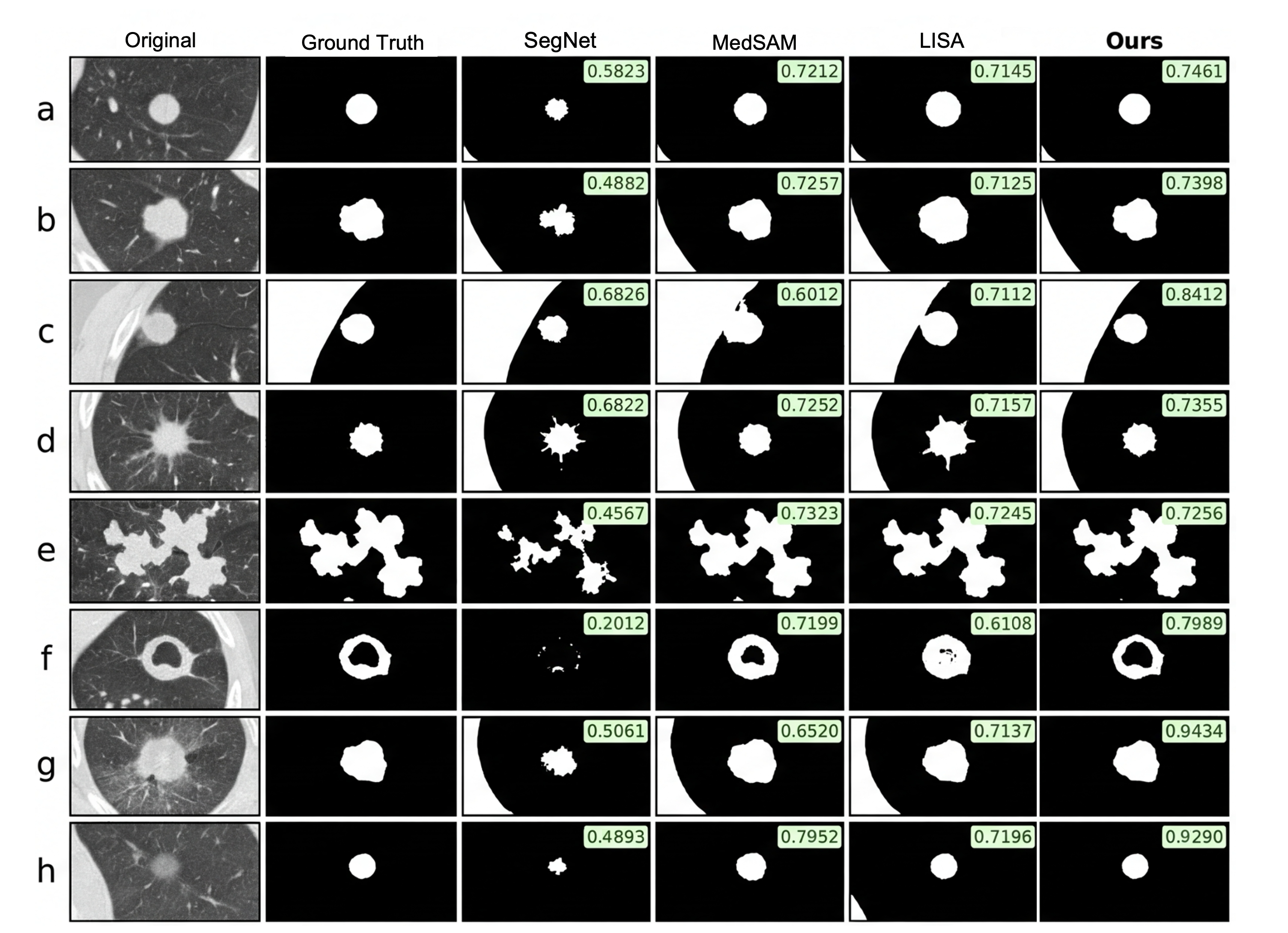}
    \caption{Qualitative comparison of segmentation results across different baseline methods. The green boxes indicate the Dice Similarity Coefficient (DSC) for each mask.}
    \label{fig:qualitative}
\end{figure*}

\textbf{Qualitative Analysis.} Visual comparisons further validate the superiority of our framework. Figure \ref{fig:qualitative} presents segmentation results across a diverse set of pulmonary lesions, including solid nodules, ground-glass opacities, and complex cavitary structures. Traditional convolutional networks such as SegNet often struggle with boundary adherence, resulting in severe under-segmentation in irregular multifocal lesions (row e). While foundation models like MedSAM and language-guided models like LISA demonstrate improved robustness, they frequently fail to capture fine-grained topological details. This limitation is visible in lesions with spiculations (row d) or internal cavities (row f). In contrast, our Semantic-Topological Graph Reasoning approach dynamically models the spatial relationships between candidate regions. This allows our framework to successfully delineate ambiguous boundaries (rows g and h) and maintain structural integrity in highly irregular pathologies, yielding predictions that closely mirror the ground truth and achieving the highest regional overlap scores consistently.

\textbf{Ablation Studies.} We incrementally integrated our core modules to isolate their contributions. Starting from a baseline LLaMA-3-V + MedSAM cascade (DSC 72.5\% on LIDC-IDRI), the introduction of Text-to-Vision Intent Distillation (TVID) provided a solid +2.8\% DSC boost by better aligning the textual prompt with visual features. Incorporating the Semantic-Topological Graph Reasoning (STGR) module yielded the most significant performance leap (+4.8\% DSC over TVID), proving its critical role in disambiguating complex anatomical overlaps. Finally, comparing fine-tuning strategies revealed a compelling insight: our SAFT approach clearly outperformed full-model tuning (DSC 81.5\% vs. 79.5\% on LIDC-IDRI). This confirms that for data-constrained medical domains, SAFT acts as a powerful regularizer, mitigating catastrophic forgetting and preventing the severe overfitting often seen when fully fine-tuning massive architectures.

\section{Discussion}
\label{sec:discussion}
Our framework establishes a robust paradigm for language-guided medical image segmentation by effectively bridging large language models and vision foundation models. The experimental results reveal several critical insights into multimodal clinical diagnostics.

\textbf{Synergy of Linguistic Priors and Visual Grounding.} Standard foundation models \cite{ma2024segment} often struggle with the low contrast of pulmonary radiographs and produce noisy predictions without precise spatial prompts. By integrating the cognitive reasoning of LLaMA-3-V through our Text-to-Vision Intent Distillation module, the model successfully translates abstract clinical terminology into concrete visual guidance. This aligns with recent findings \cite{lai2023lisa} suggesting that rich semantic embeddings significantly enhance region-level visual understanding in complex environments.

\textbf{Topological Disambiguation via Graph Reasoning.} A persistent challenge in pulmonary screening is the differentiation of overlapping anatomical structures. Traditional convolutional networks \cite{ronneberger2015u} lack the capacity to model explicit long-range spatial dependencies. By structuring candidate masks as a dynamic graph, our Semantic-Topological Graph Reasoning module effectively filters out false positives based on topological affinities. This structural awareness allows the model to maintain boundary integrity even in highly irregular pathologies.

\textbf{Mitigating Overfitting with Parameter Efficiency.} The exceptionally low variance observed across our cross-validation folds confirms the robustness of our Parameter-Efficient Fine-Tuning strategy. While full fine-tuning frequently leads to catastrophic forgetting in data-constrained medical scenarios, our Selective Asymmetric Fine-Tuning method preserves the broad knowledge of the pretrained backbones. Updating only a minimal fraction of the parameters allows the system to adapt seamlessly to specialized clinical vocabulary while avoiding severe overfitting.

\textbf{Limitations and Future Directions.} Despite the strong performance, a current limitation of our approach is its reliance on high-quality clinical text prompts provided by human experts. In real-world screening workflows, such detailed reports might not always be immediately available. Future research will focus on integrating automated report generation frameworks to create a fully autonomous diagnostic pipeline. Additionally, we plan to extend the graph reasoning module to accommodate volumetric data, enabling three-dimensional topological analysis for complex thoracic CT scans.

\section{Conclusion}
\label{sec:conclusion}
We presented a novel multimodal framework for language-guided pulmonary screening that achieves state-of-the-art segmentation performance. By integrating Text-to-Vision Intent Distillation with Semantic-Topological Graph Reasoning, our model successfully translates complex clinical instructions into precise lesion boundaries. Furthermore, our Selective Asymmetric Fine-Tuning strategy drastically reduces the computational burden of training massive foundation models while acting as a vital regularizer against overfitting. Extensive evaluations on two challenging datasets validate the superior accuracy and stability of our approach. This work provides a highly robust and resource-efficient solution for developing context-aware computer-aided diagnosis systems.

\newpage
\bibliographystyle{unsrt} 
\bibliography{reference}   

@article{DeepGBTB2026,
  title={DeepGB-TB: A Risk-Balanced Cross-Attention Gradient-Boosted Convolutional Network for Rapid, Interpretable Tuberculosis Screening},
  volume={40},
  url={https://ojs.aaai.org/index.php/AAAI/article/view/41245},
  DOI={10.1609/aaai.v40i46.41245},
  number={46},
  journal={Proceedings of the AAAI Conference on Artificial Intelligence},
  author={Lu, Zhixiang and Li, Yulong and Tang, Feilong and Jiang, Zhengyong and Li, Chong and Zhou, Mian and Li, Tenglong and Su, Jionglong},
  year={2026},
  month={Mar.},
  pages={38989--38997}
}

@inproceedings{ronneberger2015u,
  title={U-Net: Convolutional Networks for Biomedical Image Segmentation},
  author={Ronneberger, Olaf and Fischer, Philipp and Brox, Thomas},
  booktitle={Medical Image Computing and Computer-Assisted Intervention -- MICCAI 2015},
  pages={234--241},
  year={2015},
  organization={Springer}
}

@article{Lu2026skinclipvl,
  title={SkinCLIP-VL: Consistency-Aware Vision-Language Learning for Multimodal Skin Cancer Diagnosis}, 
  author={Zhixiang Lu and Shijie Xu and Kaicheng Yan and Xuyue Cai and Chong Zhang and Yulong Li and Angelos Stefanidis and Anh Nguyen and Jionglong Su},
  year={2026},
  eprint={2603.21010},
  archivePrefix={arXiv},
  primaryClass={cs.CV},
  journal={arXiv preprint arXiv:2603.21010}
}

@article{llama3,
  title={The Llama 3 Herd of Models},
  author={Llama Team},
  journal={arXiv preprint arXiv:2407.21783},
  year={2024}
}

@article{ma2024segment,
  title={Segment anything in medical images},
  author={Ma, Jun and He, Yuting and Li, Feifei and Han, Lin and You, Chenyu and Wang, Bo},
  journal={Nature Communications},
  volume={15},
  number={1},
  pages={654},
  year={2024},
  publisher={Nature Publishing Group UK London}
}

@article{Causalsamllm,
  title={Causal-SAM-LLM: Large Language Models as Causal Reasoners for Robust Medical Segmentation}, 
  author={Tao Tang and Shijie Xu and Jionglong Su and Zhixiang Lu},
  year={2026},
  eprint={2507.03585},
  archivePrefix={arXiv},
  primaryClass={cs.CV},
  journal={arXiv preprint arXiv:2507.03585}
}

@article{chen2021transunet,
  title={TransUNet: Transformers Make Strong Encoders for Medical Image Segmentation},
  author={Chen, Jieneng and Lu, Yongyi and Yu, Qihang and Luo, Xiangde and Adeli, Ehsan and Wang, Yan and Lu, Le and Yuille, Alan L and Zhou, Yuyin},
  journal={arXiv preprint arXiv:2102.04306},
  year={2021}
}

@inproceedings{liu2023grounding,
  title={Grounding DINO: Marrying DINO with Grounded Pre-Training for Open-Set Object Detection},
  author={Liu, Shilong and Zeng, Zhaoyang and Ren, Tianhe and Li, Feng and Zhang, Hao and Yang, Jie and Li, Chunyuan and Yang, Jianwei and Su, Hang and Zhu, Jun and others},
  booktitle={European Conference on Computer Vision (ECCV)},
  year={2024}
}

@article{oquab2023dinov2,
  title={Dinov2: Learning robust visual features without supervision},
  author={Oquab, Maxime and Darcet, Timoth{\'e}e and Moutakanni, Th{\'e}o and Vo, Huy and Szafraniec, Marc and Khalidov, Vasil and Fernandez, Pierre and Haziza, Daniel and Massa, Francisco and El-Nouby, Alaaeldin and others},
  journal={arXiv preprint arXiv:2304.07193},
  year={2023}
}

@inproceedings{lai2023lisa,
  title={LISA: Reasoning Segmentation via Large Language Model},
  author={Lai, Xin and Tian, Zhuotao and Chen, Yukang and Li, Yanwei and Yuan, Yuhui and Liu, Shikun and Jia, Jiaya},
  booktitle={Proceedings of the IEEE/CVF Conference on Computer Vision and Pattern Recognition (CVPR)},
  year={2024}
}

@inproceedings{zou2023segment,
  title={Segment Everything Everywhere All at Once},
  author={Zou, Xueyan and Yang, Jianwei and Zhang, Hao and Li, Feng and Li, Linjie and Gao, Jianfeng and Lee, Yong Jae},
  booktitle={Advances in Neural Information Processing Systems (NeurIPS)},
  volume={36},
  year={2023}
}

@article{Lu2026sage,
  title={SAGE: Sustainable Agent-Guided Expert-tuning for Culturally Attuned Translation in Low-Resource Southeast Asia}, 
  author={Zhixiang Lu and Chong Zhang and Yulong Li and Angelos Stefanidis and Anh Nguyen and Imran Razzak and Jionglong Su and Zhengyong Jiang},
  year={2026},
  eprint={2603.19931},
  archivePrefix={arXiv},
  primaryClass={cs.CL},
  journal={arXiv preprint arXiv:2603.19931}
}

@inproceedings{Lu2025advancing,
  title={Advancing Low-Resource Machine Translation: A Unified Data Selection and Scoring Optimization Framework},
  author={Lu, Zhixiang and Ji, Peichen and Li, Yulong and Sun, Ding and Xue, Chenyu and Xue, Haochen and Zhou, Mian and Stefanidis, Angelos and Su, Jionglong and Jiang, Zhengyong},
  booktitle={International Conference on Intelligent Computing},
  pages={482--493},
  year={2025},
  organization={Springer}
}

@article{Lu2025prism,
  title={PRISM: A Personality-Driven Multi-Agent Framework for Social Media Simulation}, 
  author={Zhixiang Lu and Xueyuan Deng and Yiran Liu and Yulong Li and Qiang Yan and Imran Razzak and Jionglong Su},
  year={2025},
  eprint={2512.19933},
  archivePrefix={arXiv},
  primaryClass={cs.CL},
  journal={arXiv preprint arXiv:2512.19933}
}

@inproceedings{dettmers2023qlora,
  title={QLoRA: Efficient Finetuning of Quantized LLMs},
  author={Dettmers, Tim and Pagnoni, Artidoro and Holtzman, Ari and Zettlemoyer, Luke},
  booktitle={Advances in Neural Information Processing Systems (NeurIPS)},
  volume={36},
  year={2023}
}

@article{Lu2026attention,
  title={Attention-based hybrid deep learning framework for modelling the compressive strength of ultra-high-performance geopolymer concrete},
  author={Xu, Minggang and Tang, Xihai and Sun, Jian and Li, Chong and Su, Jonglong and Lu, Zhixiang},
  journal={Results in Engineering},
  pages={109288},
  year={2026},
  publisher={Elsevier}
}

@inproceedings{Lu2025hierrisk,
  title={HierRisk: A Hierarchical Framework for Suicide Risk Prediction on Social Media},
  author={Lu, Zhixiang and Su, Jionglong},
  booktitle={2025 IEEE International Conference on Big Data (BigData)},
  pages={8169--8174},
  year={2025},
  organization={IEEE}
}

@inproceedings{Lu2024dieq,
  title={Dieq: Dynamic identity equilibrium for author disambiguation in kdd cup 2024 whoiswho-ind challenge},
  author={Lu, Zhixiang and Zeng, Hansheng and Li, Yuqi},
  booktitle={KDD 2024 OAG-Challenge Cup},
  year={2024}
}

@article{pedrosa2019lndblungnoduledatabase,
      title={LNDb: A Lung Nodule Database on Computed Tomography}, 
      author={João Pedrosa and Guilherme Aresta and Carlos Ferreira and Márcio Rodrigues and Patrícia Leitão and André Silva Carvalho and João Rebelo and Eduardo Negrão and Isabel Ramos and António Cunha and Aurélio Campilho},
      year={2019},
      eprint={1911.08434},
      archivePrefix={arXiv},
      primaryClass={eess.IV},
      journal={arXiv preprint arXiv:1911.08434}, 
}

@article{segnet,
      title={SegNet: A Deep Convolutional Encoder-Decoder Architecture for Image Segmentation}, 
      author={Vijay Badrinarayanan and Alex Kendall and Roberto Cipolla},
      year={2016},
      eprint={1511.00561},
      archivePrefix={arXiv},
      primaryClass={cs.CV},
      journal={arXiv preprint arXiv:1511.00561}, 
}

@article{pehrson2019automatic,
  title={Automatic pulmonary nodule detection applying deep learning or machine learning algorithms to the LIDC-IDRI database: a systematic review},
  author={Pehrson, Lea Marie and Nielsen, Michael Bachmann and Ammitzb{\o}l Lauridsen, Carsten},
  journal={Diagnostics},
  volume={9},
  number={1},
  pages={29},
  year={2019},
  publisher={MDPI}
}

@article{hu2024lung,
  title={A lung nodule segmentation model based on the transformer with multiple thresholds and coordinate attention},
  author={Hu, Tianjiao and Lan, Yihua and Zhang, Yingqi and Xu, Jiashu and Li, Shuai and Hung, Chih-Cheng},
  journal={Scientific Reports},
  volume={14},
  number={1},
  pages={31743},
  year={2024},
  publisher={Nature Publishing Group UK London}
}

@article{lu2026merit,
      title={MERIT: Multilingual Expert-Reward Informed Tuning for Chinese-Centric Low-Resource Machine Translation}, 
      author={Zhixiang Lu and Chong Zhang and Chenyu Xue and Angelos Stefanidis and Chong Li and Jionglong Su and Zhengyong Jiang},
      year={2026},
      eprint={2604.04839},
      archivePrefix={arXiv},
      primaryClass={cs.CL},
      journal={arXiv preprint arXiv:2604.04839}, 
}

\end{document}